\documentclass[10pt,twocolumn,letterpaper]{article}

\usepackage{cvpr}              

\usepackage{graphicx}
\usepackage{amsmath}
\usepackage{amssymb}
\usepackage{booktabs}
\usepackage{bm}
\usepackage{pifont}
\newcommand{\cmark}{\ding{51}}%
\newcommand{\xmark}{\ding{55}}%

\usepackage[pagebackref,breaklinks,colorlinks]{hyperref}

\usepackage[capitalize]{cleveref}
\crefname{section}{Sec.}{Secs.}
\Crefname{section}{Section}{Sections}
\Crefname{table}{Table}{Tables}
\crefname{table}{Tab.}{Tabs.}

\def\cvprPaperID{0063} 
\def\confName{CVPR}
\def\confYear{2023}

\begin{document}

\title{Making Models Shallow Again: Jointly Learning to Reduce Non-Linearity and Depth for Latency-Efficient Private Inference}

\author{Souvik Kundu\\
Intel Labs\\
San Diego, USA\\
{\tt\small souvikk.kundu@intel.com}
\and
Yuke Zhang, Dake Chen, Peter A. Beerel\\
University of Southern California\\
Los Angeles, USA\\
{\tt\small \{yukezhan, dakechen, pabeerel\}@usc.edu}
}
\maketitle

\begin{abstract}
   Large number of ReLU and MAC operations of Deep neural networks make them ill-suited for latency and compute-efficient private inference. In this paper, we present a model optimization method that allows a model to learn to be shallow. In particular, we leverage the ReLU sensitivity of a convolutional block to remove a ReLU layer and merge its succeeding and preceding convolution layers to a shallow block. Unlike existing ReLU reduction methods, our joint reduction method can yield models with improved reduction of both ReLUs  and linear operations by up to $1.73\times$ and $1.47\times$, respectively, evaluated with ResNet18 on CIFAR-100 without any significant accuracy-drop.
\end{abstract}

\section{Introduction}
\label{sec:intro}
Machine learning as a service (MLaaS) helps many users leverage the benefits of artificial intelligence (AI) augmented applications on their private data. However, due to the growing concerns associated with the model IP protection \cite{kundu2021analyzing}, the service providers often prefer to retain the model at its end rather than sharing the black box model IP with the user. Users often do not prefer sharing their personal data due to various data privacy issues. To tackle these concerns, various private inference (PI) methods \cite{mishra2020delphi,  tan2021cryptgpu, huang2022cheetah, shen2022abnn2} have been proposed that leverage techniques such as Homomorphic encryption (HE) \cite{brakerski2014efficient} and secure multi-party computation (MPC) protocols to preserve the privacy of the client's data as well as the model's IP. 
Popular PI frameworks including Gazelle \cite{juvekar2018gazelle}, DELPHI \cite{mishra2020delphi}, CryptoNAS \cite{ghodsi2020cryptonas}, and Cheetah \cite{reagen2021cheetah} leverage these privacy preserving mechanisms. However, unlike  traditional inference, the non-linear ReLU operation latency in PI can increase up to two orders of magnitude. For example, PI methods generally use Yao's Garbled Circuits (GC) \cite{yao1986generate} that demand orders of magnitude higher latency and communication than that of linear multiply-accumulate (MAC) operations.  

This has inspired the unique problem of reducing the ReLU operations and thus, the latency overhead associated with PI. In particular, earlier works leveraged neural architecture search \cite{ghodsi2020cryptonas, cho2021sphynx} to  reduce the ReLU layer count. Recent works \cite{cho2022selective, kundu2023learning} proposed model linearization via systematic ReLU reduction that can reduce ReLU at the level of layer, channel, and pixel. However, all these works focus on reducing \textit{only} the ReLU operation. Interestingly, the reduced ReLU models may often have comparable total non-linear and total linear MAC operation latency. Moreover, recent improvement of non-linearity operation latency in PI via oblivious transfer \cite{huang2022cheetah} has essentially demanded simultaneous reduction of ReLU as well as MAC operations.   

\begin{figure}[t]
    \centering
    \includegraphics[width=0.82\columnwidth]{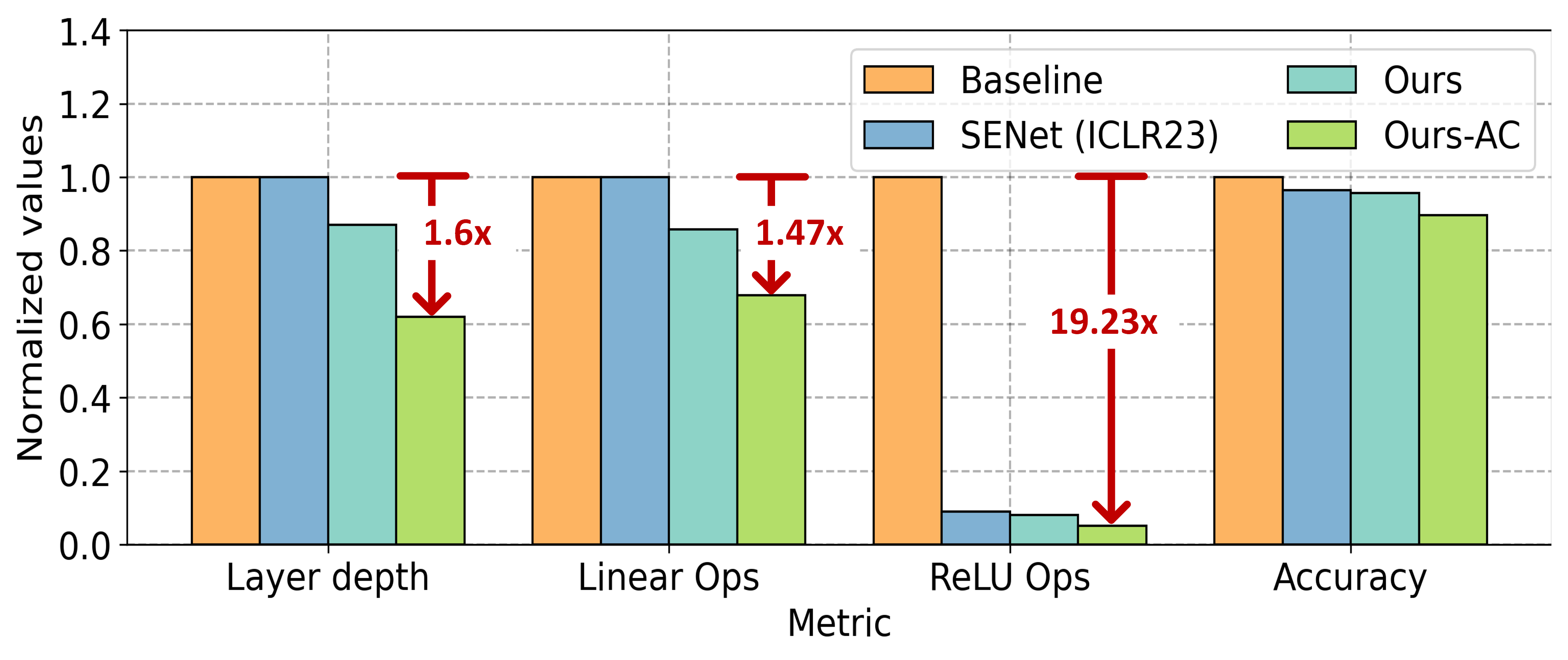}
    \vspace{-2mm}
    \caption{Comparison of the proposed method with the baseline and the existing state-of-the-art (SoTA) \cite{kundu2023learning}. Compared to existing SoTA, our method reduces both the linear and ReLU operations of the model at the cost of in-significant accuracy drop. We normalized each metric with respect to the maximum value of that metric. ``Ours-AC" represents the model with AC output. We used ResNet18 on CIFAR-100 for evaluation.}
    \vspace{-5mm}
    \label{fig:intro_comp}
\end{figure}

\textbf{Our contributions.} Towards this goal, we present a model architecture optimization framework that simultaneously learns to reduce both the ReLUs and the MACs of a model. In particular, we systematically replace every convolutional block (example: basic-block layer for ResNet18) having a low ReLU budget \cite{kundu2023learning}, with a shallower block having no non-linearity module allowing us to reduce both MACs and ReLUs. We term this method as \textit{gated branching} (GB) as we allow gradual learning of the branch with a shallower block via a gating condition, that changes over epochs (refer to Fig. \ref{fig:overall_framework}(b)).  To further reduce the depth at the later layers, we present a \textit{auxiliary knowledge distillation} (AKD) to an auxiliary classifier (AC) at a shallower depth of the DNN model. We conduct experiments with ResNet18 on CIFAR-10, CIFAR-100, and WRN-22-8 on CIFAR-100 datasets to evaluate the efficacy of the proposed framework. As Fig. \ref{fig:intro_comp} shows, our joint learning method can yield linear and ReLU operation reduction of up to $1.47\times$ and $19.23\times$ compared to the baseline models.

\section{Priliminaries}
\label{sec:prelim}
\textbf{Private inference.} We assume a semi-honest client-server PI scenario where a client, holding private data, intends to use inference service from a server having a private model \cite{mishra2020delphi}. Here, each party tries to reveal their collaborator's private data by inspecting the information they received while strictly following the protocol. To protect from data revelation, we assume the PI to happen on encrypted data \cite{mishra2020delphi, ghodsi2020cryptonas, lou2020safenet} in an offline-online format \cite{mishra2020delphi}, that transfers input data independent computations to the offline stage while performing the data-dependent operations online.

\textbf{ReLU reduction for efficient PI.} Existing works use model designing with reduced ReLU counts via methods including the search for ReLU-efficient models \cite{mishra2020delphi, ghodsi2020cryptonas, lou2020safenet, cho2021sphynx} and manual ReLU-importance-driven non-linearity reduction \cite{jha2021deepreduce}. More recently, \cite{cho2022selective} leveraged $l_1$-regularization to remove ReLU at the granularity of both pixel and channels to yield improved accuracy vs. non-linearity trade-off. Finally, a contemporary work \cite{kundu2023learning} demonstrated an automated ReLU sensitivity evaluation method and for a given ReLU budget, leveraged that to propose a three-stage training framework, dubbed as SENet, towards linearizing a model via ReLU reduction. In particular, for a given ReLU budget, at stage 1, SENet automatically identifies per-layer ReLU count. During stage 2, it learns the binary mask associated to the ReLU locations in a non-linear layer of the partial ReLU model (PR) model. 1 value in a mask position assigns a ReLU unit for that position while the 0 value assigns an identity unit. Finally, during stage 3, it fine-tunes the PR model with the mask frozen. SENet yields the SoTA accuracy vs. non-linearity trade-off, and can meet a target ReLU budget without any costly hyper-parameter tuning that is necessary for the $l_1$-regularization method \cite{cho2022selective}. For these notable advantages, we adopt SENet's training method into our optimization framework.

\section{Motivational Analysis}
\label{sec:motive}
\begin{figure}[t]
    \centering
    \includegraphics[width=0.83\columnwidth]{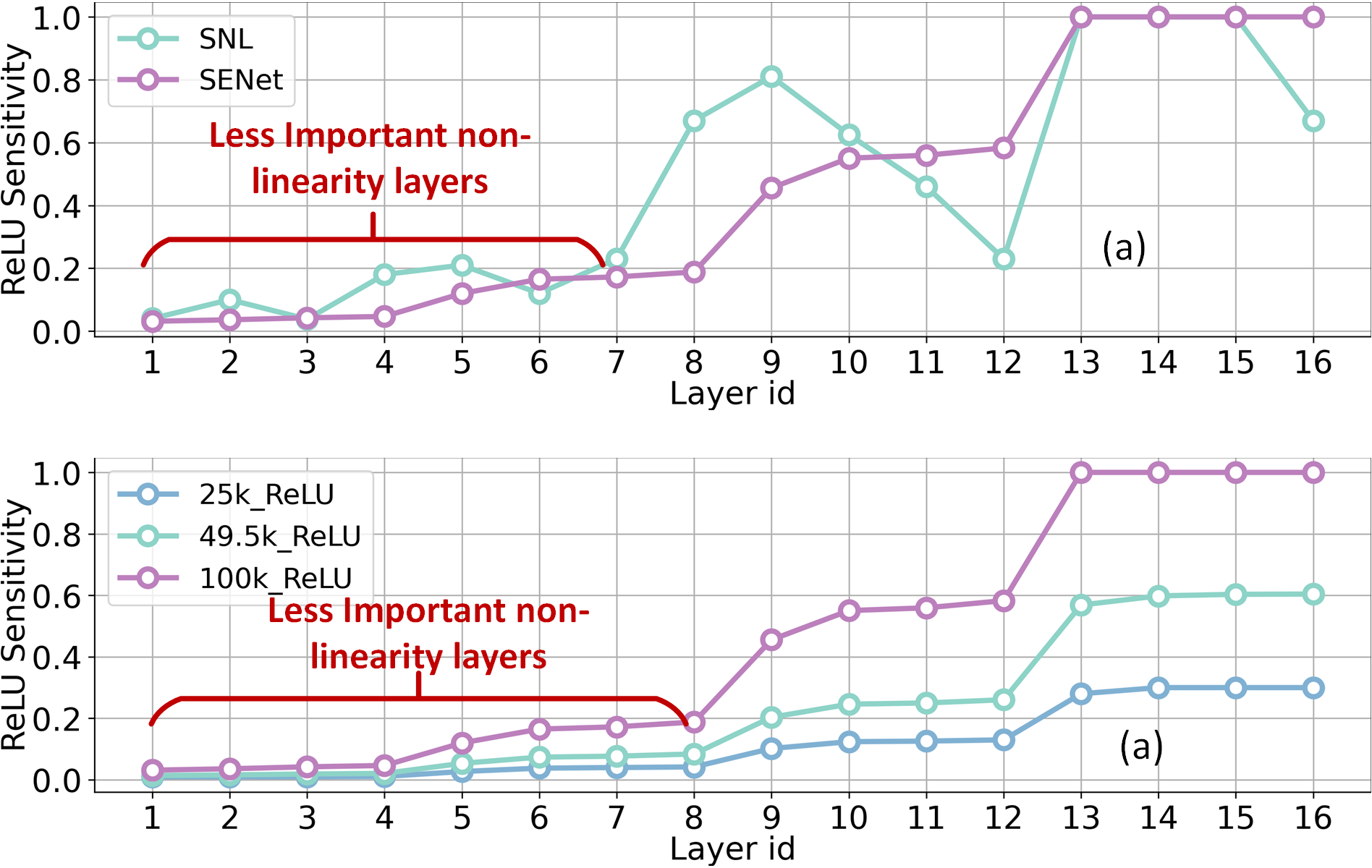}
    \vspace{-2mm}
    \caption{(a) Comparison of layer-wise ReLU sensitivity evaluated via SNL\cite{cho2022selective} and SENet \cite{kundu2023learning} for a ReLU budget of $100$k; (b) Comparison of layer-wise ReLU sensitivity for different ReLU budgets using SENet training method. We used the basic block ReLUs of a ResNet18 trained on CIFAR-100 for both the plots.}
    \vspace{-5mm}
    \label{fig:motivation_sens}
\end{figure}

For a given ReLU budget, we start by evaluating the ReLU sensitivity\footnote{ReLU sensitivity of a layer is measured as the ratio of ReLUs assigned in a layer (mask location value of 1) to the total number of possible ReLU units of that layer.} of the non-linearity layers of a DNN. To perform this analysis, we used a ResNet18 model and trained it on CIFAR-100 using two recent automated ReLU reduction frameworks, SNL \cite{cho2022selective} and SENet \cite{kundu2023learning}. We keep a target ReLU budget of $100$k for both. As demonstrated in Fig. \ref{fig:motivation_sens}(a), both the frameworks yield ReLU distribution sensitivity with the initial layers having less sensitivity than the later layers. As we understand, high ReLU sensitivity may be co-related to high non-linearity importance \cite{kundu2023learning}, the earlier non-linearity layers may play a less significant role in retaining the classification performance as compared to the later non-linearity layers.

We further conducted ReLU sensitivity analysis for three different ReLU budgets of $25$k, $49.5$k, and $100$k with the ResNet18 on CIFAR-100. As demonstrated in Fig. \ref{fig:motivation_sens}(b), the earlier non-linearity layers consistently show less importance compared to the later ones, for all the three ReLU budgets. Inspired by these observations, we pose the following intriguing question. 

\textit{Is it possible to design a shallow DNN by removing the less important ReLU layers while retaining the classification performance?}

\section{Methodology}
\label{sec:method}

\begin{figure*}[t]
    \centering
    \includegraphics[width=1.75\columnwidth]{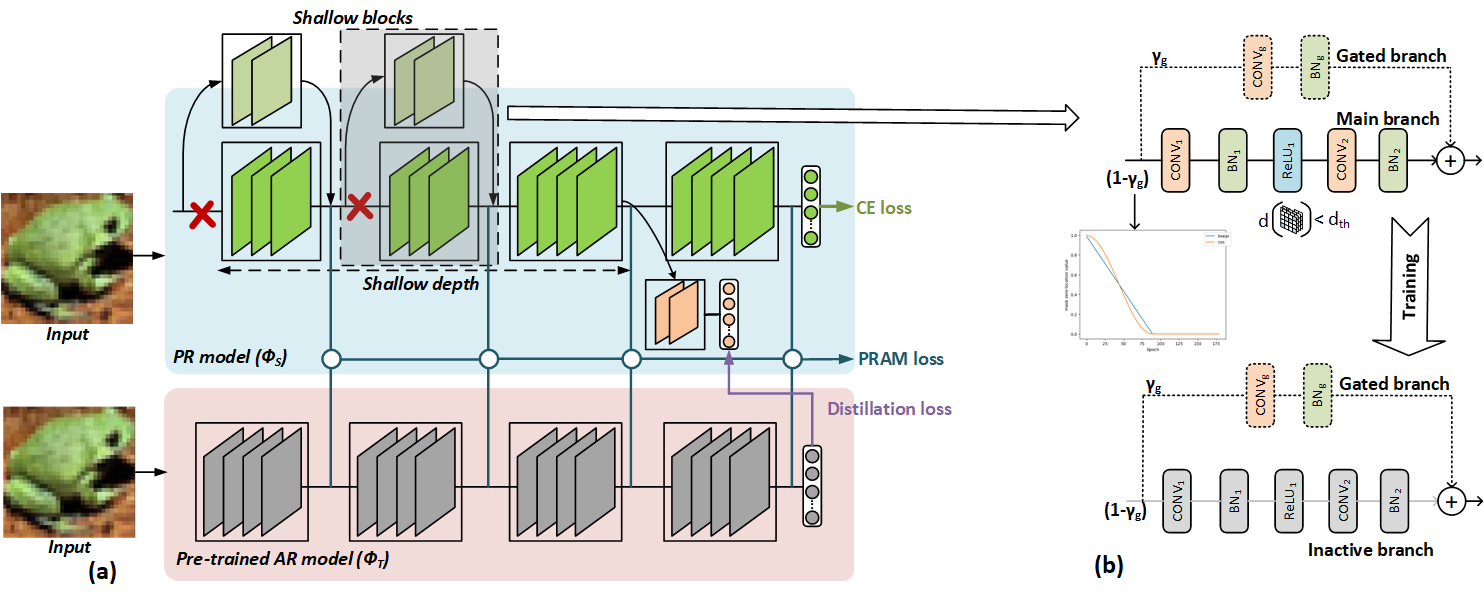}
    \vspace{-2mm}
    \caption{(a) Fine-tuning framework of the partial ReLU (PR) model that uses gated branching to reduce the depth of the earlier blocks, while learning a shallow auxiliary classifier to  reduce the depth for later blocks. (b) The gradual training procedure to learn the gated branching path when the non-linearity budget of the main branch is below a certain threshold.}
    \vspace{-5mm}
    \label{fig:overall_framework}
\end{figure*}

We now describe our training framework that can jointly reduce the ReLU and the depth of a DNN. While we follow \cite{kundu2023learning}, to train a partial ReLU model via a three-stage optimization framework, we replace the stage 3 of \cite{kundu2023learning} with our novel fine-tuning stage. In the first stage, we identify the per-layer ReLU count for a given ReLU budget by computing the normalized ReLU sensitivity \cite{kundu2023learning}. For a layer $l$, the ReLU sensitivity $\eta_{\bm{\alpha}^l}$ can be computed as \cite{kundu2023learning},
\begin{align}
    \eta_{\bm{\alpha}^l} = (1 - \eta_{\bm{\theta}^l}),
\end{align}
\noindent
Where, $\eta_{\bm{\theta}^l}$ represents the pruning sensitivity \cite{kundu2021attentionlite, kundu2021dnr, kundu2022towards, kundu2022sparse} of the preceding layer represented in terms of the fraction of non-zero weights for a given parameter density. Next, in stage 2, we compute a binary mask $M^l \in \{0, 1\}^{h^l\times w^l \times c^l}$ for each non-linearity layer $l$, where $h^l, w^l$, and $c^l$ represents the height, width, and the number of channels in the activation map. The 1's in $M^l$ represents the locations with ReLU units while 0's represents identity units in the activation tensor.  During the fine-tuning stage (stage 3), we present a training strategy for joint reduction of ReLU operation and depth of the PR model.

 In the our fine-tuning stage, we first introduce a threshold ReLU sensitivity value denoted by $d_{th}$, that represents the minimum ReLU sensitivity for a non-linearity layer to be retained. In particular, at the end of stage 3, for a convolutional layer block (example, basic block for ResNets), we only keep those ReLU layers that have a sensitivity $d > d_{th}$. Removal of a non-linear layer effectively connects the preceding and succeeding linear layers (example convolution) one after the other. It is well known that a function represented through a sequence of linear function layers like convolution, can be approximated with a single linear function layer \cite{fu2022depthshrinker}. We thus replace the convolution batch-normalization (CONV-BN) pairs located immediately before and after the dropped ReLU layer, with a single CONV-BN unit. 

\textbf{Gated branching.} As during fine-tuning, we initialize the PR model weights with that of the best model evaluated during stage 2, modifying the model architecture with shallow layer blocks (as depicted in Fig.  \ref{fig:overall_framework}(a) at the earlier blocks of $\Phi_S$) with untrained weights is not straight forward. In particular, direct replacement of some of the blocks with shallow blocks from the beginning of fine-tuning stage may significantly reduce the PR model's final performance due to sub-optimal training. We thus present a gated branching (GB) method that assigns a shallow branch for the layer blocks having ReLU sensitivity $<d_{th}$ (see Fig. \ref{fig:overall_framework}(b)). We assign a gating hyperparameter $\gamma_g$ to the shallow branch with the main branch having the weight factor of $(1-\gamma_g)$. Thus, for an input tensor $\bm{X}$, the functional representation of the combined block is $\gamma_g(f_{GB}(\bm{X})) + (1 - \gamma_g)(f_{MB}(\bm{X}))$, with $f_{GB}$ and $f_{MB}$ representing the functions represented by the gated and the main branch, respectively. We start the $\gamma_g$ from 0 and gradually increase it to 1 with the training epochs using either a cosine or linear increment function. This allows a gradual introduction of the shallow branches helping the PR model to transfer knowledge from the corresponding deeper blocks that finally become inactive at the end of the fine-tuning. In particular, we gradually increase $\gamma_g$ to 1 till the end of the epoch when the first learning-rate (LR) decay occurs. We keep $\gamma_g$ fixed to 1 during the remaining part of the fine-tuning allowing the fixed architecture with shallow branches to fine-tune.
Note, we keep the first ReLU after the CONV stem, out of the GB method.

\textbf{Auxiliary knowledge distillation.} It is note-worthy that reducing the depth via GB, may be applicable for a DNN's initial layers only, due to the less importance of non-linearity in those layers. Later non-linearity layers, on the other hand, play a critical role in retaining the classification performance, and thus ReLU layer removal via the thresholding method may cost a huge accuracy drop. To reduce depth at the later layers, we present an auxiliary classifier (AC) branch located after a shallow depth of the DNN. In particular, inspired by the model architecture used in self-distillation \cite{zhang2019your}, we place the AC before the final convolutional layer block (refer to Fig. \ref{fig:overall_framework}(a)). This architecture allows the PR model ($\Phi_S$) to learn and predict at two layer depths, one at the original classifier and the other at AC at a shallower depth. To further improve the accuracy of $\Phi_S$, we present an \textit{auxiliary knowledge distillation} (AKD) that distills the knowledge from an all-ReLU (AR) baseline model ($\Phi_T$) to the AC of the $\Phi_S$.

\begin{table*}[!t]
\caption{Comparison between the proposed method with gated branching (w/ GB) and SoTA reduction method \cite{kundu2023learning} (w/o GB). $\uparrow$ and $\downarrow$ indicate the higher the better and the lower the better, respectively.}
\begin{center}
\scriptsize\addtolength{\tabcolsep}{-0.0pt}
\begin{tabular}{c|c|c|c|c|c|c|c|c|c|c|c|c}
\hline
Model & Baseline  & $d_{th}$ & \multicolumn{2}{c|}{\#ReLUs (k) $\downarrow$} & \multicolumn{2}{c|}{Accuracy (\%) $\uparrow$}  & \multicolumn{2}{c|}{Layer depth $\downarrow$}    & \multicolumn{2}{c}{MAC saving $\uparrow$} & \multicolumn{2}{|c}{ReLU ops reduction $\uparrow$} \\
\cline{4-13}
 & Acc (\%) &  & w/o GB & w/ GB &  w/o GB & w/ GB  & w/o GB & w/ GB & w/o GB & w/ GB & w/o GB & w/ GB\\
\hline
\multicolumn{13}{c}{\textcolor{black}{Dataset: CIFAR-10}}\\
\hline
ResNet18 & 95.2 & 0.1 & 82 & \textbf{76.8} & 93.05 & \textbf{93.05}& 16 & \textbf{13}  & $1\times$ & $\textbf{1.17}\times$ & $6.8\times$ & $\textbf{7.3}\times$ \\
\hline
\multicolumn{13}{c}{\textcolor{black}{Dataset: CIFAR-100}}\\
\hline
ResNet18 & 78.05 & 0.05 & 24.9 & \textbf{21.1} & \textbf{70.59} & 69.10 & 16 & \textbf{12}  & $1\times$ & $\textbf{1.3}\times$ & $21.8\times$ & $\textbf{26.4}\times$ \\ 
\hline
ResNet18 & 78.05 & 0.05 & 49.6 & \textbf{47.4} & \textbf{75.28} & 74.62 & 16 & \textbf{14}  & $1\times$ & $\textbf{1.16}\times$ & $11.2\times$ & $\textbf{11.8}\times$ \\ 
\hline
WRN22-8 & 80.82 & 0.1 & 240 & \textbf{221} & \textbf{79.81} & 79.62 & 18 & \textbf{15} & $1\times$ & $\textbf{1.15}\times$ & $5.8\times$ & $\textbf{6.3}\times$ \\
\hline
\end{tabular}
\end{center}
\label{tab:comaprison}
\vspace{-4mm}
\end{table*}

We thus, combine the two methods of GB and AKD, allowing additional ReLU reduction both at the initial and later layers during the fine-tuning stage, compared to that proposed in SENet \cite{kundu2023learning}. Moreover, the systematic depth reduction makes the model yielded by our method have significantly fewer linear operations compared to the SoTA alternatives. let $\Psi_{pr}^m$ and $\Psi_{ar}^m$ represent the $m^{th}$ pair of vectorized post-ReLU activation maps of same layer for $\Phi_{pr}$ and $\Phi_{ar}$, respectively. Our loss function for the fine-tuning phase is given by
\begin{align}\label{eq:sp_distill_loss}
        {\mathcal{L}} =  & \lambda\underbrace{{\mathcal{L}}_{KL}\left(\sigma\left(\frac{z^{ar}}{\rho}\right),\sigma\left(\frac{z^{pr}_{ac}}{\rho}\right)\right)}_{\text{KL-div. loss}} + (1-\lambda)\underbrace{{\mathcal{L}}_{pr}(y,y^{pr})}_{\text{CE loss}}\nonumber\\
    & + \frac{\beta}{2}\sum_{m \in I}\underbrace{\left\lVert\frac{\Psi_{pr}^m}{\lVert \Psi_{pr}^m\rVert_2} - \frac{\Psi_{ar}^m}{\lVert \Psi_{ar}^m\rVert_2}\right\lVert_2}_{\text{PRAM loss}}.
\end{align}
\noindent
where $\sigma$ is the softmax function with $\rho$ being KL-div. temperature. $\lambda$ balances the importance between the CE and KL divergence loss components, and $\beta$ is the weight for the post-ReLU AR-PR activation mismatch (PRAM) loss.

\section{Experiments}
\label{sec:expt}
\subsection{Experimental Setup}
To evaluate the efficacy of our method, we performed experiments with ResNet18 \cite{he2016deep}and  wide residual network 22-8 (WRN22-8) \cite{zagoruyko2016wide}. In particular, we evaluated ResNet18 on two popular datasets, CIFAR-10 and CIFAR-100 \cite{krizhevsky2009learning} and WRN22-8 on CIFAR-100. We used PyTorch API to define and train our models on an Nvidia RTX 2080 Ti GPU. 

To perform stage 1 and 2, we followed the same hyperparameters as \cite{kundu2023learning}. To perform joint ReLU and depth reduction during fine-tuning stage, we initialized a PR model with the weights that provided the  best accuracy during the mask evaluation stage (stage 2). We then trained the model for 180 epochs with starting LR of 0.01 that decayed by a factor of 10 at epochs 90, 140, and 160. Unless otherwise stated, we used a linear increase of $\gamma_g$ each epoch by $\frac{i}{90}$ and kept it to $1$ for $i > 90$, where $i$ represents the epoch number. Similar to \cite{kundu2023learning}, we weighted the loss of each component by setting $\lambda = 0.9$, $\beta=1000$, and used KD temp. $\rho=4$.

\begin{table}[!t]
\caption{Performance of our method with depth reduction both at the earlier layers via GB and later layers via AC output from an intermediate shallow layer. The AC is placed after the $3^{rd}$ basic-block layer of a ResNet18. `\cmark' and `\xmark' in AC output indicates whether output is taken at the AC or the final classifier.}
\begin{center}
\scriptsize\addtolength{\tabcolsep}{-3.5pt}
\begin{tabular}{c|c|c|c|c|c|c|c}
\hline
Model & Baseline & With GB  & AC   & Accuracy  & ReLU ops & MAC & Layer \\
 & Acc (\%) &  & output &  (\%) $\uparrow$ & reduction $\uparrow$ & saving $\uparrow$ & depth $\downarrow$\\
\hline
\multicolumn{7}{c}{\textcolor{black}{Dataset: CIFAR-10}}\\
\hline
ResNet18 & 95.2 & \cmark & \xmark & 95.17 & $3.94\times$ & $1.15\times$ & 14\\
         &      & \cmark & \cmark & 94.51 & $5.07\times$ & $1.46\times$ & 10 \\
\hline
\multicolumn{7}{c}{\textcolor{black}{Dataset: CIFAR-100}}\\
\hline
ResNet18 & 78.05 & \cmark & \xmark & 78.24 & $5.8\times$ & $1.15\times$ & 14\\
         &      & \cmark & \cmark & 76.31 & $8.5\times$ & $1.46\times$ & 10 \\
\hline
\end{tabular}
\end{center}
\label{tab:gb_ac_results}
\vspace{-4mm}
\end{table}
\subsection{Model Performance Analysis}
\textbf{Results with GB.} Table \ref{tab:comaprison} demonstrates results of the PR models with GB. Note, for models with GB, similar to \cite{kundu2023learning}. Due to the shallow layer blocks, the models with GB can provide MAC saving and reduce the layer depth by up to $1.3\times$ and $1.33\times$, respectively, while yielding  reduction of ReLU operation by up to $26.4\times$. To train models with GB, apart from the CE loss of $\Phi_S$, we used the PRAM loss and a KL-div. between the final classifiers of $\Phi_S$ and $\Phi_T$. 

\textbf{Results with GB and AKD.} Table \ref{tab:gb_ac_results} demonstrates the results of the models whose depth are reduced via both GB and AKD with the final classification results taken from the AC of the PR model. In particular, at the cost of modest accuracy drop, the AC can provide an additional depth reduction of $1.4\times$ compared to the one using only GB. Further, classification via AC reduces both the ReLU and MAC operations by up to $1.46\times$ and $1.27\times$, respectively, than models using only GB.

\section{Conclusions}
\label{sec:conc}
We introduced a training method to jointly reduce both the ReLU operations and layer depth of a DNN model suitable for latency-compute-efficient PI. In particular, we present a gated branching method to learn a shallow block in replacing the deep CONV layer blocks having non-linearity layers with low ReLU sensitivity. We further place an auxiliary classifier to allow classification at a shallower depth compared to that with original classifier. Empirical demonstration showed that the shallow-reduced-ReLU models can yield improved reduction of both ReLUs  and layer depth by up to $1.73\times$ and $1.6\times$, respectively, without any significant accuracy-drop, compared to that yielded via SoTA ReLU reduction methods.
{\small
\bibliographystyle{ieee_fullname}
\bibliography{egbib}
}

\end{document}